\documentclass[lettersize,journal]{IEEEtran}
\usepackage{amsmath,amsfonts}
\usepackage{algorithmic}
\usepackage{algorithm}
\usepackage{array}
\usepackage[caption=false,font=normalsize,labelfont=sf,textfont=sf]{subfig}
\usepackage{textcomp}
\usepackage{stfloats}
\usepackage{url}
\usepackage{multirow} 
\usepackage{booktabs} 
\usepackage{verbatim}
\usepackage{graphicx}
\usepackage{cite}
\begin{document}

\title{Single-Cell Deep Clustering Method Assisted by Exogenous Gene Information: A Novel Approach to Identifying Cell Types}
\author{Dayu~Hu, Ke~Liang, Hao~Yu and~Xinwang~Liu$^{\dagger}$,~\IEEEmembership{Senior~Member,~IEEE}
\thanks{Dayu Hu,  Ke Liang, Hao Yu, Xinwang Liu are with the School of Computer, National University of Defense Technology, Changsha, China, 410073. Email: hzauhdy@gmail.com, liangke200694@126.com, csyuhao@gmail.com, xinwangliu@nudt.edu.cn.}
\thanks{$^{\dagger}$ Corresponding author.}}

\markboth{Journal of \LaTeX\ Class Files,~Vol.~14, No.~8, August~2021}%
{Shell \MakeLowercase{\textit{et al.}}: A Sample Article Using IEEEtran.cls for IEEE Journals}


\maketitle

\begin{abstract}
In recent years, the field of single-cell data analysis has seen a marked advancement in the development of clustering methods.  Despite advancements, most of these algorithms still concentrate on analyzing the provided single-cell matrix data.  However, in medical applications, single-cell data often involves a wealth of exogenous information, including gene networks.  Overlooking this aspect could lead to information loss and clustering results devoid of significant clinical relevance.  An innovative single-cell deep clustering method, incorporating exogenous gene information, has been proposed to overcome this limitation.  This model leverages exogenous gene network information to facilitate the clustering process, generating discriminative representations. Specifically, we have developed an attention-enhanced graph autoencoder, which is designed to efficiently capture the topological features between cells. Concurrently, we conducted a random walk on an exogenous Protein-Protein Interaction (PPI) network, thereby acquiring the gene's topological features. Ultimately, during the clustering process, we integrated both sets of information and reconstructed the features of both cells and genes to generate a discriminative representation. Extensive experiments have validated the effectiveness of our proposed method. This research offers enhanced insights into the characteristics and distribution of cells, thereby laying the groundwork for early diagnosis and treatment of diseases.
\end{abstract}
\begin{IEEEkeywords}
Exogenous gene information, Clustering, Protein-protein interaction, Node2vec, Deep learning.
\end{IEEEkeywords}

\section{Introduction}
\label{sec:introduction}

\IEEEPARstart{S}{ingle-cell}  transcriptome sequencing technology represents a significant advancement in the field of genomics. It elucidates the intricate biological processes at the cellular level and serves as a potent tool for studying the origins and microenvironments of tumors \cite{sun2021tisch,peng2022cell,zhang2020single,hu2023integrating}. Unsupervised clustering represents a pivotal step in this process. By analyzing the gene expression data of individual cells, it precisely differentiates between various cell types and states. This approach provides valuable insights into understanding complex biological systems, such as cancer, neurodegenerative diseases, and developmental processes. However, owing to the complexity of biological systems, devising a clustering algorithm that is both accurate and highly clinically relevant remains a formidable challenge.

\begin{figure}[!t]%
\centering
\includegraphics[width=1\linewidth]{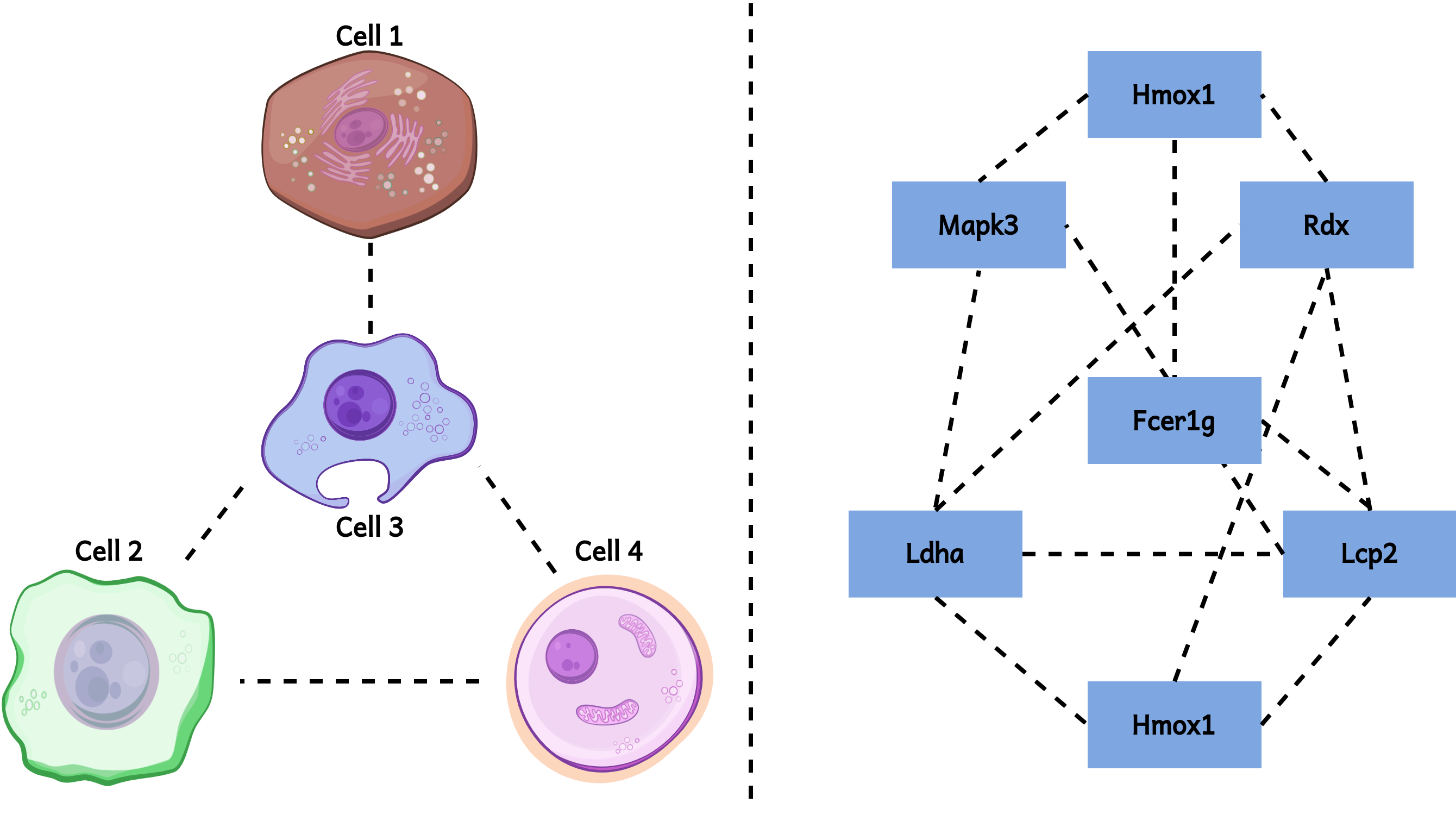}
\caption{Cells and genes both exhibit associative relationships.   The left illustrates the connections between cells, while the right depicts the associations among genes.}\label{fig1}
\end{figure}

In recent years, a significant increase in the development of clustering algorithms tailored for single-cell RNA sequencing (scRNA-seq) data has been observed\cite{liu2023deep,yan2023bmvae}. Early approaches depended on probabilistic models that estimated high-dimensional cell data through computing the probability of gene expression. For instance, CIDR\cite{lin2017cidr} introduced an interpolation method to handle dropout events, whereas SC3\cite{kiselev2017sc3} employed hierarchical $k$-means clustering to facilitate consensus clustering, presuming Euclidean relationships between cells. However, these methods operate under the assumption that biological data are linear and clear, an assumption that may not always be valid in the real world.

To effectively extract features from scRNA-seq data and circumvent assumptions about data distribution, some researchers have suggested neural networks as a promising approach for mining information from scRNA-seq data\cite{mao2023gene,liu2023cross}. Neural networks, widely used as black box models, can adapt to nearly all data distributions when the parameters are appropriately configured. Numerous single-cell deep clustering algorithms have been proposed to obtain effective representations, A detailed introduction to these methods will be provided in the Related Work section (Section 2A). However, these models often treat cells as isolated entities, overlooking the associations between them.

 To integrate cellular interaction relationships into the clustering process, researchers have proposed graph-based approaches for deriving cell embeddings. This approach necessitates constructing a cell graph based on intercellular similarities. These constructed graphs, in conjunction with the original feature matrix, are subsequently inputted into a graph neural network for training. A detailed introduction to graph clustering algorithms will be provided in Section 2B. Although these graph-based deep clustering algorithms have progressed in capturing the topological features of cells, their focus remains primarily on analyzing the provided single-cell matrix data.  However, clustering algorithms oriented towards medical applications should integrate external information for a more holistic analysis, as overlooking this aspect could result in clustering outcomes that diverge from clinical conclusions.

Single-cell data inherently contain exogenous information.  Unlike other datasets, the features in scRNA datasets are meaningful as they represent genes.  Biologists and medical scientists have extensively explored gene relationships.  Despite this extensive research, most current clustering methods still overlook these gene relationships, focusing solely on cell connections and neglecting gene associations.  However, in reality, genes within each cell participate in complex interrelations due to interactions, regulatory mechanisms, and shared functions and pathways in biological processes.  In essence, as is shown in Figure \ref{fig1}, gene topological features are present, yet this aspect remains largely unexplored in single-cell clustering research.  By extracting and integrating the topological features of gene interconnectivity into the clustering framework, significant optimization of clustering embeddings and enhancement of clustering outcomes can be achieved.  Furthermore, this type of embedding could lead to a more accurate representation of biological characteristics, thereby enhancing the alignment between identified clusters and the actual underlying biological systems.

In light of these considerations, we have developed an exogenous gene information-assisted single-cell deep clustering method (scEGA) that simultaneously focuses on the interaction relationships between cells and genes.To accomplish this, we utilized a graph attention autoencoder (GAT), which captures the topological structure between cells and ensures effective information transmission among them.   Additionally, we conducted random walks on the exogenous protein-protein interaction (PPI) network corresponding to the gene set, to obtain embeddings that represent the gene's topological features. During the clustering process, we integrated these two elements and reconstructed the features of both cells and genes, thereby acquiring a discriminative cell representation. Experiments on eight real scRNA datasets demonstrate that our scEGA method is stable and outperforms nine other baseline methods in performance.   Our contributions can be summarized as follows:

\begin{itemize}
\item The scEGA model simultaneously focuses on the cell features and exogenous gene features, fusing and aligning them during the clustering process to generate a more discriminative representation.

\item The scEGA model employs a dual-supervised module to facilitate the optimization of the bottleneck layer, effectively utilizing its own information and requiring no labels.

\item The scEGA model is robust and demonstrates superior performance compared to the other nine baseline methods.
\end{itemize}

\section{Related work}
\subsection{Single-cell Deep Clustering}

Recently, deep learning methods have been widely applied to analyze scRNA-seq data due to their formidable learning capabilities.  Li et al. proposed DESC, which iteratively learns the gene expression pattern of each cluster, assigns cells to their respective clusters and continuously mitigates batch effects \cite{li2020deep}.  Tian et al. propose scDeepCluster \cite{tian2019clustering}, a method rooted in the Zero-inflated Negative Binomial (ZINB) model, utilizing a bottleneck layer for deep $k$-means clustering to enhance clustering outcomes.  Tian et al. developed a deep embedding clustering approach for single-cell data (scDCC), integrating the ZINB model with clustering loss and constraint loss \cite{tian2021model}.  However, these deep neural networks struggle to preserve the topological structure of scRNA-seq data, because they neglect the associations between cells during analysis.

\subsection{Single-cell Deep Graph Clustering}

The advent of deep graph autoencoders has addressed the aforementioned concerns, namely, that previous models treated cells as isolated individuals. These graph autoencoders efficiently learn cluster-friendly, low-dimensional representations by incorporating graph topology information of cell-to-cell interactions. Satija et al. proposed Seurat \cite{satija2015spatial}, which employs Louvain community detection to construct a cell graph, subsequently analyzed through spectral clustering using Phenograph. Wang et al. proposed scGNN \cite{wang2021scgnn}, which utilizes a graph neural network to capture and integrate relationships between cells, complemented by a Gaussian model to represent the pattern of heterogeneous gene expression.  Yu et al. introduced scTAG \cite{yu2022zinb}, a specialized deep graph embedding clustering algorithm tailored for single-cell data, which concurrently optimizes clustering loss, ZINB loss, and cell graph reconstruction loss. Furthermore, Chen proposed scGAC \cite{cheng2022scgac}, which introduces attention mechanisms based on the cell-to-cell graph, thus ensuring effective information transmission between cells.   Meanwhile, our previous model, scDFC \cite{hu2023scdfc}, combines structural data from cell-to-cell graphs with attribute information from cellular expression patterns, thereby facilitating a comprehensive analysis of scRNA data.

\section{Methods}

\subsection{Preliminary}
Single-cell data refer to genetic expression information obtained through single-cell sequencing technology, which is presented in matrix form. In this work, we provide a simple mathematical description of this data, represented as a numerical matrix denoted by $\mathbf{X}\in \mathbb{R}^{N \times D}$, where $D$ denotes the dimension of genes, and $N$ represents the number of cells.

\begin{figure*}[!t]%
\centering
\includegraphics[width=0.9\linewidth]{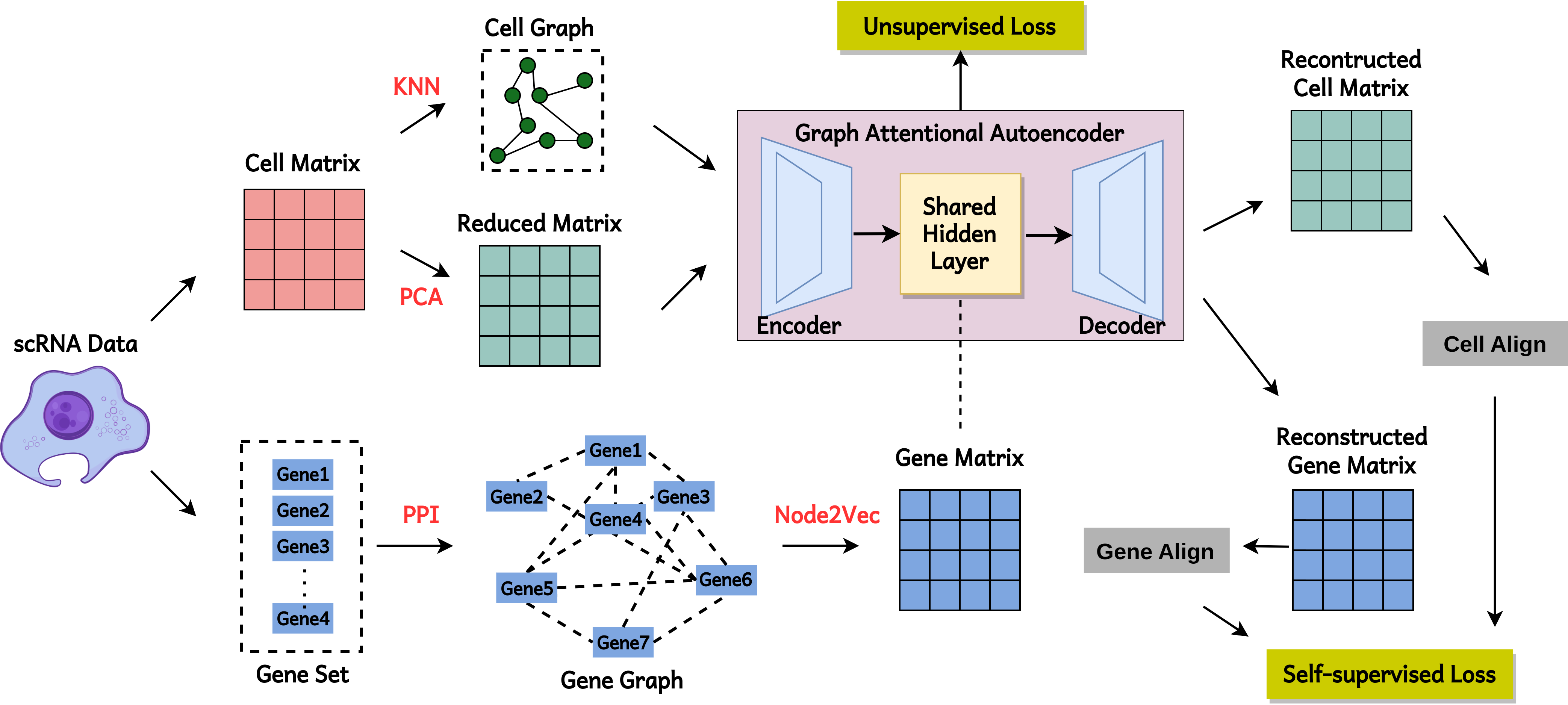}
\caption{An illustration of the scEGA model framework. This framework primarily consists of two modules: a dual matrix alignment module, which allows gene representations to participate in the optimization process of deep clustering, thereby fully utilizing the exogenous information of single-cell datasets. The second is a dual-supervised optimization module, which effectively optimizes embeddings through self-supervised and unsupervised loss.}\label{fig2}
\end{figure*}

\subsection{The Framework of scEGA}\label{subsec2}

Figure \ref{fig2} depicts the comprehensive workflow of the scEGA model, which consists of two main modules. The first module is the dual-matrix alignment module. This module processes the cell dataset and gene set in parallel to construct two separate matrices: the reconstructed cell matrix and the reconstructed gene matrix. First, to construct the cell-to-cell graph, a k-nearest neighbors (KNN) approach based on similarity measures is employed, and then a specific adjacency matrix $\mathbf{A}$ is obtained. This adjacency matrix, along with the reduced cell matrix, is then fed into a graph attentional autoencoder, as shown in Figure \ref{fig2}. It outputs the reconstructed cell matrix. On the other hand, we also construct the gene-to-gene graph, using exogenous information. Specifically, the gene set is input into the online STRING website to generate a PPI network corresponding to the gene set. This network is then fed as the input of the node2vec algorithm to perform random walks and obtain the final reconstructed gene matrix.

The second module of the model is the dual-supervised optimization module. This module benefits from two supervised mechanisms and relies on no external labels. Initially, the reconstructed matrices of the cell and the gene are obtained. Then, a self-supervised mechanism is used to constrain the stability of these two matrices during the embedding optimization process. Additionally, the shared bottleneck layer is optimized with an unsupervised mechanism to ensure that the embedding exhibits exceptional clustering performance.

\subsection{Dual-matrix Alignment Module}
This section provides a detailed description of the dual-matrix fusion module, including the definition of formulas for dual-matrix graphs and the specific computational procedure.

\subsubsection{Cell Matrix}
To accurately learn the cell-to-cell graph information, we designed a graph-based autoencoder (GAT) enhanced by an attention mechanism to fully capture the cell signaling patterns and cell-to-cell relationships. The original feature matrix $\mathbf{X}$, after undergoing principal component analysis (PCA) reduction, yields the dimensionally reduced matrix. The reduced matrix $\mathbf{\hat{X}}$ is encoded to produce the cell embedding, obtained with the following equation:
\begin{equation}
\mathbf{H}_{c}=\sigma\left(\mathbf{W}_{e}^{c} \mathbf{A}\mathbf{\hat{X}}\right),
\end{equation}

\noindent the encoding weight parameter matrix $\mathbf{W}_{e}^{c} $ in the Graph Attention Network (GAT) consists of learnable parameters that map the input features to the bottleneck layer. During training, each element of $\mathbf{W}_{e}^{c} $ is subject to adjustment. The nonlinear activation function $\sigma$ facilitates the neural network's ability to learn complex patterns and features.

Subsequently, the embedding $\mathbf{H}_{c}$, derived from the GAT, is integrated with the embedding produced by the gene graph, which is denoted as $\mathbf{H}_{g}$. This fusion process is executed as follows:
\begin{equation}
\mathbf{H}_{fusion}=\sigma\left([\mathbf{H}_{c}|| \mathbf{H}_{g}]\right),
\end{equation}

\noindent the shared embedding of the two graphs is represented as $\mathbf{H}_{fusion}$, and the concatenation operation is represented using $||$. The nonlinear activation function $\sigma$ remains the same as previously described. Subsequently, the decoding module reconstructs the shared embedding, expressible as:
\begin{equation}
\mathbf{X}_{r}=\mathbf{W}_{d}^{c} 
\mathbf{H}_{fusion},
\end{equation}

\noindent where $\mathbf{W}_{d}^{c}$ is the learnable decoding matrix of GAT.

\subsubsection{Gene Matrix}\label{subsubsec2}

\begin{figure}[!t]%
\centering
\includegraphics[width=0.7\linewidth]{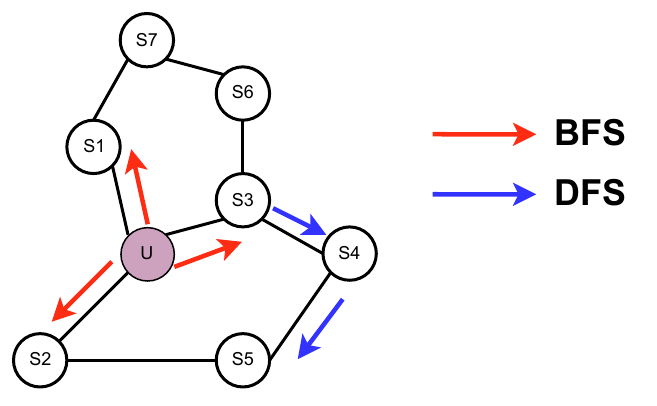}
\caption{In BFS and DFS traversals, the node pointed to by the red line is considered a low-order neighbor of the source node, while the node pointed to by the blue line is considered a higher-order neighbor.}\label{fig3}
\end{figure}

This section offers a comprehensive guide for constructing a gene-to-gene graph from a single-cell dataset. Initially, we processed each single-cell dataset by utilizing Scanpy to identify highly variable genes. We retained the top 2000 genes as the final gene set. Subsequently, the gene set was uploaded to the online platform STRING\footnote{https://www.string-db.org/} to generate a PPI network. The network was then saved as an adjacency table to facilitate the subsequent random walk. To generate node embeddings for the PPI network, we utilized the biased approach node2vec for random walk, involving two neighborhood strategies: breadth-first search (BFS) and depth-first search (DFS). BFS focuses on traversing nodes of the same order, while DFS emphasizes traversing higher-order nodes. Figure \ref{fig3} depicts the detailed procedure of the random walks. By employing these two strategies, node2vec ensures more effective walks, leading to enhanced node embeddings.

Formally, consider $\mathbf{G} = (\mathbf{V}, \mathbf{E})$ as a PPI network, with $\mathbf{V}$ representing the set of nodes, each corresponding to a protein, and $\mathbf{E}$ indicating the interactions between these proteins. We utilize node2vec to generate an embedding vector for each node. Specifically, the sequence $\mathbf{S}_{v}$ is treated as a corpus, employing the skip-gram model to capture the features. The objective of the skip-gram model is to maximize the probability of observing node $v$ within a specific context. Consequently, for node $v$, the maximization of the following likelihood function is pursued:
\begin{equation}
\frac{1}{|\mathbf{S}_{v}|}\sum_{u \in \mathbf{S}_{v}} \sum_{j \in \mathcal{N}_u} \log \mathbf{P}(v_j | v_u),
\end{equation}

\noindent in this context, $\mathcal{N}_u$ represents the set of neighboring nodes of node $u$, while $\mathbf{P}(v_j | v_u)$ defines the conditional probability of node $j$ given node $u$. The calculation of this probability employs the softmax function, defined as follows:
\begin{equation}
\mathbf{P}(v_j | v_u) = \frac{\exp(v_j^\top v_u)}{\sum{k \in V} \exp(v_k^\top v_u)},
\end{equation}

\noindent here, $v_u$ and $v_j$ represent the embedding vectors of nodes $u$ and $j$, respectively.  It is assumed that the PPI network comprises ${n}$ nodes.  The gene embedding $\mathbf{Z}_{g}$ is derived as follows:
\begin{equation}
\mathbf{Z}_{g}=[v_1,v_2,\ldots,v_n]^\top,
\end{equation}

\noindent the gene embedding $\mathbf{Z}_{g}$ from the PPI network are inputted into a neural network for joint training and subsequently reconstructed via a decoder.  The training process is characterized as follows:
\begin{equation}
\mathbf{H}_{g}=\sigma\left(\mathbf{W}_{e}^g\mathbf{Z}_{g}\right),
\end{equation}

\noindent where $\mathbf{W}_{e}^g$ is the learnable encoding matrix of gene-to-gene graph, and $\sigma$ is the nonlinear activation function. In a similar manner, the reconstructed gene matrix $\mathbf{Z}_{g}^r$ is computed utilizing the equation below:
\begin{equation}
\mathbf{Z}_{g}^r=\mathbf{W}_{d}^g
\mathbf{H}_{fusion}
\end{equation}

\noindent where $\mathbf{W}_{d}^g$ denotes the learnable decoding matrix for the gene-to-gene graph.

\subsection{Dual-Supervised Optimization Module}
\subsubsection{Self-supervised Optimization}

The aim of self-supervised optimization is to utilize the intrinsic features inherent in the data.  In the scEGA framework, the self-supervised loss was employed to ensure the stability of both the cell and gene matrices during the embedding optimization process.  This was specifically achieved by aligning the original input data with its reconstructed counterpart.  For the cell matrix, the objective was to obtain a reconstructed version that closely resembles the original matrix.  This was accomplished by employing cosine loss, a widely-used similarity metric, to regulate similarity during the clustering process, as demonstrated below:
\begin{equation}
\mathcal{L}_{\mathrm{cell}} = \frac{\mathbf{\hat{X}} \cdot \mathbf{X}_r}{\|\mathbf{\hat{X}}\| \cdot \|\mathbf{X}_r\|}.
\end{equation}

Regarding the gene matrix, the aim was to maintain the reconstructed gene data unaltered. To accomplish this, Mean Absolute Error (MAE) loss was utilized to guarantee the functional integrity of the genes throughout the clustering optimization process. The particular alignment process unfolds as follows:
\begin{equation}
\mathcal{L}_{\mathrm{gene}}  = |\mathbf{Z}_{g}-\mathbf{Z}_{g}^r|,
\end{equation}

\noindent the final self-supervised loss is combined as below:

\begin{equation}
\mathcal{L}_{\mathrm{ssl}}={2\lambda \mathcal{L}_{\mathrm{cell}}+2(1-\lambda)\mathcal{L}_{\mathrm{gene}}},
\label{eq11}
\end{equation}

\noindent where $\lambda$ represents a tunable hyperparameter.

\subsubsection{Unsupervised Optimization}

In this study, the student's t-distribution was utilized to optimize the bottleneck layer. The matrix $\mathbf{Q}$ encapsulates the cluster assignments for each cell, as illustrated below:
\begin{equation}
q_{i j}=\frac{\left(1+\left\|z_{i}-u_{j}\right\|^{2}\right)^{-1}}{\sum_j\left(1+\left\|z_{i}-u_{j}\right\|^{2}\right)^{-1}},
\end{equation}

\noindent where $z$ represents the embedding of a cell, and $u$ signifies the center of clustering. Subsequently, an auxiliary target distribution $\mathbf{P}$ was constructed based on the clustering distribution $\mathbf{Q}$.
\begin{equation}
p_{i j}=\frac{q_{i j}^{2} / \sum_i q_{i j}}{\sum_j\left(q_{i j}^{2} / \sum_i q_{i j}\right)},
\end{equation}

\noindent The objective of optimizing $\mathbf{Q}$ is to closely approximate it to $\mathbf{P}$. To this end, the Kullback-Leibler (KL) divergence was employed as a constraint, termed the unsupervised loss, denoted as follows:
\begin{equation}
\mathcal{L}_{\mathrm{ul}}=\sum_{i}\sum_{j} p_{i j} \log \frac{p_{i j}}{q_{i j}}.
\label{eq16}
\end{equation}

In summary, the dual-supervised optimization module utilizes a two-step learning approach, comprising self-supervised optimization grounded in the data's inherent characteristics, and unsupervised optimization anchored in the student's t-distribution. Self-supervised optimization is realized via dual-matrix alignment, preserving the consistency of cell matrix and gene matrix throughout clustering, thereby facilitating the learning of a significant compressed representation. Unsupervised optimization seeks to minimize the discrepancy between the actual clustering assignment and the auxiliary target clustering distribution, resulting in embeddings characterized by superior clustering performance. By integrating these two optimization approaches, the total loss  of our model $\mathcal{L}_{\mathrm{f}}$ is formulated as follows:
\begin{equation}
\mathcal{L}_{\mathrm{f}}= \mathcal{L}_{\mathrm{ssl}}+\mathcal{L}_{\mathrm{ul}}.
\end{equation}

\begin{table}[tbp]
    \vspace{-10pt}
     \renewcommand{\arraystretch}{1.2}
    \caption{Details of the eight real datasets.}
    \label{tab1} 
    \centering
    \resizebox{0.4\textwidth}{!}{
        \begin{tabular}{cccc}
            \toprule
            Datasets & Genes & Clusters & References \\
            \midrule
            Biase    & 21489 & 3 & \cite{biase2014cell} \\
            Darmanis & 9337  & 8 & \cite{darmanis2015survey} \\
            Enge     & 25929 & 9 & \cite{enge2017single} \\
            Bjorklund & 26087 & 4 & \cite{bjorklund2016heterogeneity} \\
            Sun      & 995   & 6 & \cite{sun2019bayesian} \\
            Marques  & 15291 & 14 & \cite{marques2018transcriptional} \\
            Zeisel   & 18825 & 9 & \cite{zeisel2015cell} \\
            Fink     & 20932 & 7 & \cite{fink2022single} \\
            \bottomrule
        \end{tabular}
    }
    \vspace{-10pt}
\end{table}

\section{Experiments}
In this section, extensive experiments were conducted to evaluate our model.  The subsequent sections will cover Experimental Settings, Clustering Performance, Ablation Study, and Parameter Fine-Tuning.

\begin{table*}[!htbp]
	\caption{ARI scores of scEGA and baseline methods across eight datasets, with the top three results highlighted in bold. '-' indicates the error of the method itself.}
	\label{tab2}
	\centering
 \renewcommand{\arraystretch}{1.2}
	\resizebox{1\textwidth}{!}{
		\begin{tabular}{|c|c|c|c|c|c|c|c|c|c|c|}
			\hline
			\multirow{2}{*}{Datasets} & \multicolumn{2}{c|}{Traditional Methods} 
	& \multicolumn{3}{c|}{Deep Clustering Methods} & \multicolumn{5}{c|}{Deep Graph Clustering Methods}\\
	\cline{2-11}
	& CIDR & SC3 & scDeepCluster & DESC &scAE & scGNN & Seurat &  scGAC & scDFC & \textbf{scEGA}  \\

\hline
			Biase       & \textbf{1.000} & 0.948 & 0.948 & 0.960 & \textbf{1.000}& 0.330 & 0.850  & \textbf{1.000} & \textbf{1.000} & \textbf{1.000} \\
			Darmanis    & 0.337 & 0.470 & 0.522 & \textbf{0.536} & 0.100& - & 0.353  & 0.508 & \textbf{0.533} & \textbf{0.549} \\
			Enge        & 0.223 & \textbf{0.531} & 0.218 & 0.305 & 0.052& - & 0.206  & 0.261 & \textbf{0.368} & \textbf{0.483} \\
			Bjorklund   & 0.457 & 0.721 & 0.310 & 0.412 & - & 0.438 & 0.056 & \textbf{0.785} & \textbf{0.842} & \textbf{0.724} \\
			Sun         & 0.268 & \textbf{0.879} & 0.783 & 0.603 & 0.276& 0.465 & 0.182  & \textbf{0.784} & \textbf{0.784} & \textbf{0.834} \\
			Marques     & 0.100 & \textbf{0.363} & \textbf{0.390} & 0.269& - & - & 0.172  & 0.283 & 0.260 & \textbf{0.399} \\
			Zeisel      & 0.167 & 0.420 & \textbf{0.736} & 0.279 & -& - & 0.119  & \textbf{0.657} & 0.317 & \textbf{0.605} \\
			Fink        & 0.225 & 0.146 & \textbf{0.359} & 0.212 & 0.179& - & 0.063  & 0.328 & \textbf{0.485} & \textbf{0.561} \\
			\hline
		\end{tabular}
	}
\end{table*}

\subsection{Experimental Settings}
\subsubsection{Datasets}
This research presents eight real-world datasets from prevalent species, including humans and mice. To assess the efficacy of various clustering algorithms, each dataset is supplemented with definitive labels. The following provides a succinct introduction to each dataset:

\begin{itemize}
\item \textbf{Biase}\cite{biase2014cell}, a representative small scRNA dataset, primarily comprises embryonic cells derived from mice, collected during their developmental stages.
\end{itemize}

\begin{itemize}
\item \textbf{Darmanis}\cite{darmanis2015survey} includes human brain cells, known for their complex composition, leading to its division into multiple clusters.
\end{itemize}

\begin{itemize}
\item \textbf{Enge}\cite{enge2017single} is comprised of pancreatic cells from humans.
\end{itemize}
\begin{itemize}
\item \textbf{Bjroklund}\cite{bjorklund2016heterogeneity} encompasses lymphoid cells from humans, crucial to the immune system.
\end{itemize}

\begin{itemize}
\item \textbf{Sun}\cite{sun2019bayesian} offers three single-cell datasets, with this study focusing on the first one, containing exclusively mouse lung cells.
\end{itemize}

\begin{itemize}
\item \textbf{Marques}\cite{marques2018transcriptional} includes data from mice, aimed at investigating the developmental origin of oligodendrocyte precursor cells.
\end{itemize}

\begin{itemize}
\item \textbf{Zeisel}\cite{zeisel2015cell} contains data from mice, sourced from the somatosensory cortex and hippocampus CA1 regions.
\end{itemize}

\begin{itemize}
\item \textbf{Fink}\cite{fink2022single}, sourced from the human adult ureter, may provide insights into metabolic processes.
\end{itemize}

The initial scRNA data exhibit significant variability in scale and high noise levels, which could potentially lead to erroneous conclusions in subsequent analyses. To mitigate these issues, quality control was conducted on the cellular data prior to clustering. Specifically, cells with expression values within a reasonable range were retained, and outliers with extreme expression values were eliminated. This was achieved by establishing upper and lower thresholds at 75$\%$ plus three times the quartile deviation, and 25$\%$ minus the quartile deviation, respectively. Following quality control, the data were standardized by scaling to a consistent range. Subsequently, a log2 transformation was applied to the data. To avoid negative infinite values and ensure positive expression values, a pseudo count of 1 was incorporated during the transformation process.

\subsubsection{Compared Methods}
This section offers a concise overview of the baseline methods employed in these experiments.
\begin{itemize}
\item \textbf{CIDR}\cite{lin2017cidr} utilizes a probabilistic model for evaluating dropout events in cellular data, categorized as a traditional clustering method in this study.
\end{itemize}

\begin{itemize}
\item \textbf{SC3}\cite{kiselev2017sc3} implements a consensus clustering approach using $k$-means clustering and Euclidean distance, identified as a traditional clustering method in this study.

\end{itemize}

\begin{itemize}
\item \textbf{scDeepCluster}\cite{tian2019clustering} introduces a deep autoencoder using the ZINB loss, classified as a deep clustering method in this research.
\end{itemize}

\begin{itemize}
\item \textbf{DESC}\cite{li2020deep} employs an autoencoder network for cell embedding and batch effect elimination, distinguished as a deep clustering method in this research.
\end{itemize}

\begin{itemize}
\item \textbf{scGNN}\cite{wang2021scgnn} combines three iterative multimodal autoencoders based on graph neural networks, recognized as a deep graph clustering method in this study.
\end{itemize}

\begin{itemize}
\item \textbf{Seurat}\cite{satija2015spatial} features a built-in Phenograph clustering method for constructing cell graphs via community detection, identified as a graph clustering method in this research.
\end{itemize}

\begin{itemize}
\item \textbf{scAE} represents a simple deep clustering model constructed for comparative analysis, classified as a deep clustering method in this study.

\end{itemize}

\begin{itemize}
\item \textbf{scGAC}\cite{cheng2022scgac} introduces an attention mechanism in graph neural networks for efficient cellular graph construction, distinguished as a graph deep clustering method in this research.
\end{itemize}

\begin{itemize}
\item \textbf{scDFC}\cite{hu2023scdfc} merges cell attribute information with structural inter-cell information for clustering, recognized as a deep fusion clustering method in this study.
\end{itemize}

\subsubsection{Implementation Details}
The performance of the proposed algorithm was evaluated on an Ubuntu server featuring an Intel Core i7-6800K CPU, 64GB of DDR4 memory, and an NVIDIA TITAN Xp graphics card.  The system utilized Ubuntu 22.04.2 LTS, and the algorithm was implemented in Python 3.6, using TensorFlow deep learning framework version 1.12.0.  The parameters of the random walk on the gene graph in the scEGA model were set to the default parameters of the node2vec algorithm.  Encode layer sizes were set to (512, 256, 64), with the bottleneck layer established at 64.  The model underwent pre-training for 200 epochs, followed by a training phase lasting 5000 epochs.  Learning rates were set at 0.0002 for pre-training and 0.0005 for the training phase.

\subsubsection{Evaluation}

This study utilizes three evaluation metrics, detailed as follows:

\begin{itemize}
\item
\textbf{Adjusted Rand Index (ARI)}\cite{hubert1985comparing} is a widely utilized metric for measuring the consistency between clustering results and true labels, necessitating labeled data. The formulation of this index is as follows:
\end{itemize}
\begin{equation}
\label{eq:21}
\begin{aligned}
\text{ARI} = \frac{\sum_{ij}\binom{n_{ij}}{2}-[\sum_{i}\binom{a_{i}}{2}\sum_{j}\binom{b_{j}}{2}]/\binom{n}{2}}{\frac{1}{2}[\sum_{i}\binom{a_{i}}{2}+\sum_{j}\binom{b_{j}}{2}]-[\sum_{i}\binom{a_{i}}{2}\sum_{j}\binom{b_{j}}{2}]/\binom{n}{2}}.
\end{aligned}
\end{equation}

\begin{itemize}
\item \textbf{Normalized Mutual Information (NMI)}\cite{strehl2002cluster} is another commonly used metric to assess the similarity between clustering results and true labels, requiring labeled data. The formulation of this index is as follows:
\end{itemize}
\begin{equation}
\label{eq:16}
\begin{aligned}
\text{NMI}=\frac{2 M I(U, V)}{H(U)+H(V)}.
\end{aligned}
\end{equation}

\begin{itemize}
\item \textbf{Silhouette Coefficient (SC)}\cite{rousseeuw1987silhouettes} serves as an internal evaluation metric for assessing clustering quality in an unsupervised manner, eliminating the need for ground truth labels. SC ranges from -1 to 1, where a value of 1 signifies high compactness and distinct separation between clusters, -1 signifies low compactness and poor separation, and 0 indicates overlapping clusters. Contrary to the aforementioned metrics, SC offers a more holistic evaluation by considering both within-cluster and between-cluster distances. In this study, SC is utilized to control the early stopping criterion in our Python implementation.
\end{itemize}
\begin{equation}
\text{SC}=\frac{b_{i}-a_{i}}{\max \left(a_{i}, b_{i}\right)}.
\end{equation}

\subsection{Clustering Performance}

\begin{figure}[!t]%
\centering
\includegraphics[width=1\linewidth]{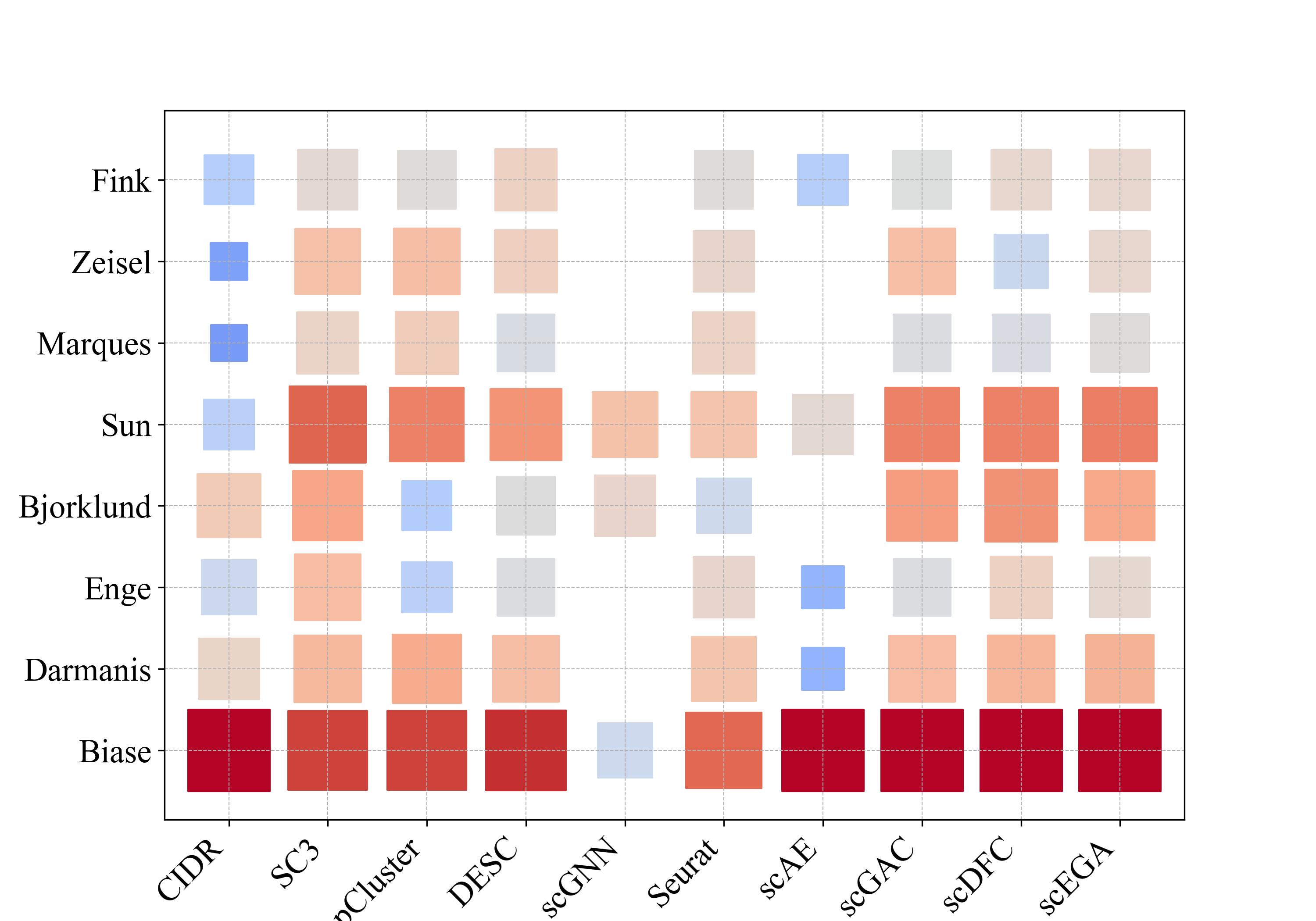}
\caption{The heatmap presents the clustering performance of scEGA as evaluated by NMI.  The size and color intensity of each square in the heatmap correspond to the NMI values: larger and darker squares indicate higher NMI values.}\label{fig4}
\end{figure}

We carried out a comprehensive series of experiments to assess the effectiveness of established benchmark clustering methods, encompassing our proposed scEGA model and nine other baseline methods. The findings unequivocally indicate that scEGA consistently achieved superior performance in the ARI, as detailed in Table \ref{tab2}. scEGA consistently ranked within the top three in all comparative analyses and secured the best in half of the datasets (four out of eight). 

To visually illustrate these findings, we constructed a heatmap based on NMI for depicting clustering performance, as shown in Figure \ref{fig4}. Each square's size in the heatmap is indicative of the NMI values' magnitude, with a gradation in color from lighter to darker shades to represent ascending NMI values. It is noteworthy that models such as CIDR, scDeepCluster, and scAE displayed comparatively lower clustering performance in NMI terms. Conversely, scEGA not only illustrated its dominance in ARI but also its significant stability in NMI. Given the complexity of the biological environment, which contributes to the intricate distribution of single-cell data, identifying a universally applicable clustering method poses a considerable challenge. Nonetheless, scEGA displayed high performance in nearly all the evaluated tasks.

\begin{figure*}[!t]%
\centering
\includegraphics[width= 1\linewidth]{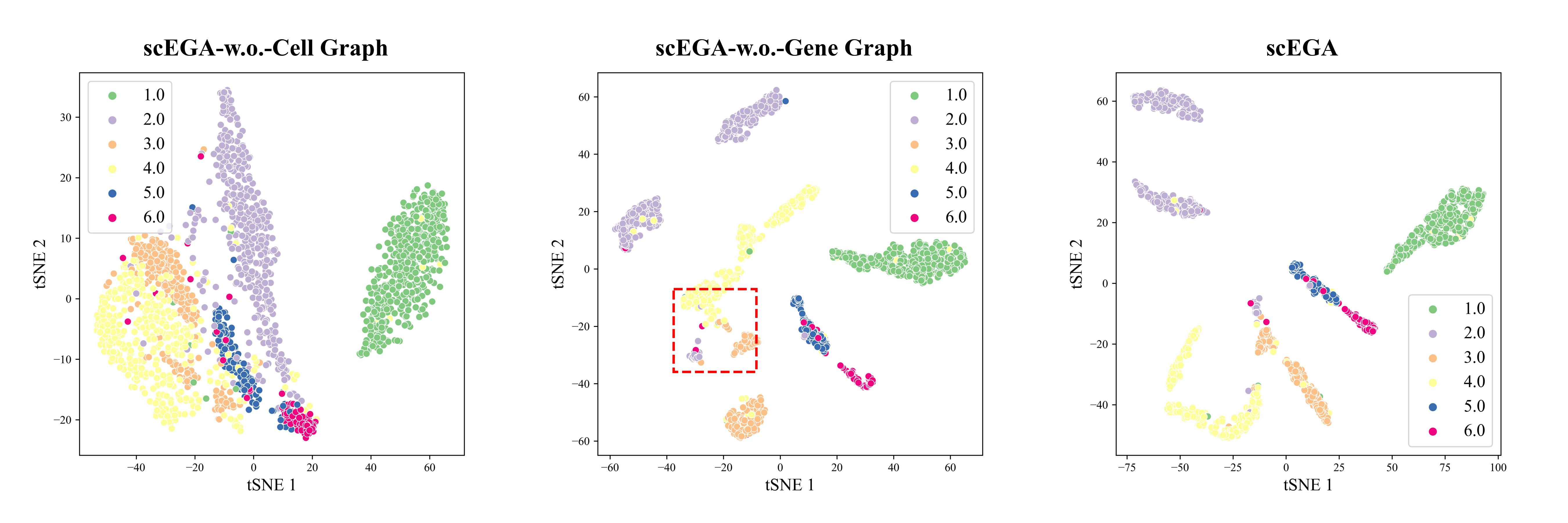}
\caption{The visualization illustrates different embeddings on the Sun dataset, resulting from the absence of certain modules in scEMC: scEGA 'w.o.' (without)  the cell graph, scEGA 'w.o.' the gene graph, and scEGA inclusive of all modules. }\label{fig5}
\end{figure*}

\begin{table}[tbp]
    \vspace{-10pt}
      \renewcommand{\arraystretch}{1.2}
    \caption{The ablation study utilizing ARI values: We successively removed the cell graph and gene graph from the model to observe the respective impacts on performance.}
    \label{tab3} 
    \centering
    \resizebox{0.5\textwidth}{!}{
        \begin{tabular}{cccc}
            \toprule
            Dataset & scEGA-w.o.-Cell Graph & scEGA-w.o.-Gene Graph & scEGA \\
            \midrule
            Biase     & 0.983 & 1.000 & \textbf{1.000} \\
            Darmanis  & 0.443 & 0.500 & \textbf{0.549} \\
            Enge      & 0.077 & 0.189 & \textbf{0.483} \\
            Bjorklund & -     & 0.513 & \textbf{0.724} \\
            Sun1      & 0.282 & 0.736 & \textbf{0.834} \\
            Marques   & 0.235 & -     & \textbf{0.399} \\
            Zeisel    & 0.246 & 0.293 & \textbf{0.605} \\
            Fink      & 0.110 & -     & \textbf{0.561} \\
            \bottomrule
        \end{tabular}
    }
    \vspace{-10pt}
\end{table}

\subsection{Ablation Study}

\subsubsection{Quantitative Analysis}
The dual-matrix alignment module, central to our scEGA model, hinges on the integration of the cell graph and gene graph. To investigate these modules' impact on clustering outcomes, we performed ablation experiments. Specifically, we developed two scEGA variants, one without the cell graph module and the other without the gene graph module. We evaluated their clustering performance in comparison to the complete scEGA model. As indicated in Table \ref{tab3}, scEGA demonstrated superior clustering performance. Removing the cell graph module led to a significant decrease in performance, underscoring the cell-to-cell network's importance and its pivotal role in clustering. While the removal of the gene map module did not cause as substantial a decline in clustering performance, the noticeable decrease still emphasized the exogenous gene features' vital role in clustering. The integration of both graphs yielded the best performance, suggesting that our dual-matrix alignment module effectively and reliably enhances clustering.

\subsubsection{Visualization Analysis}
The quality of cell embeddings directly influences clustering performance.   In this section, we present a series of visualization analyses on the embeddings of various scEGA variants.   Specifically, we performed experiments by separately removing the cell graph module and the gene graph module from the model.   We then saved the bottleneck layer and visualized it using t-SNE on the Sun dataset.   Our findings indicate a significant decline in the quality of embeddings in the absence of these modules.   As illustrated in Figure \ref{fig5} (left), the omission of the cell graph led to poor clustering, thereby hindering the ability to differentiate between clusters.  Highlighted in the red box in Figure \ref{fig5} (mid), removing the gene graph caused a subset of cells to be incorrectly assigned to clusters.   These findings underscore the critical role of both the cell graph and gene graph modules in scEGA in achieving high-quality cluster embeddings.

\begin{figure}[!t]%
\centering
\includegraphics[width=1\linewidth]{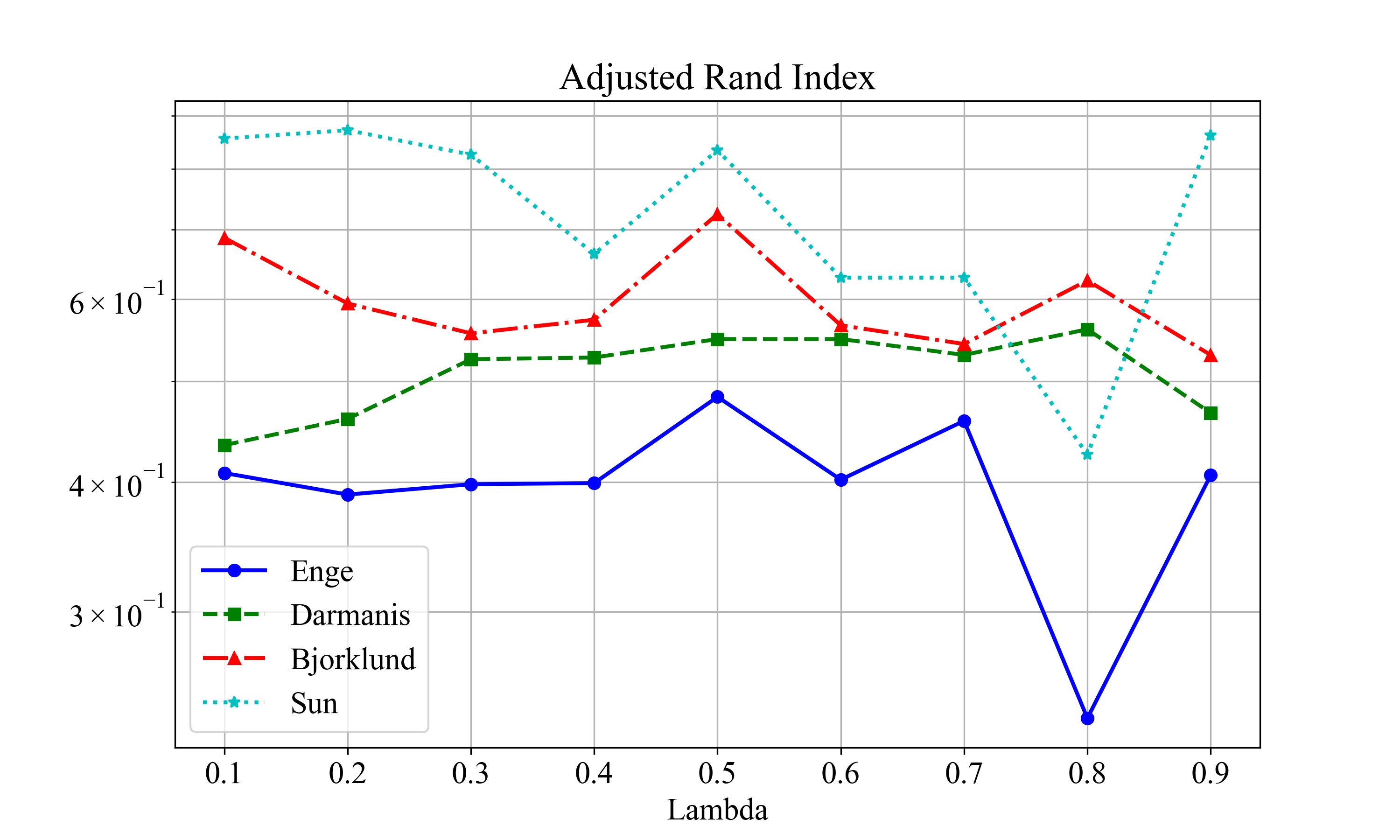}
\caption{Investigating optimal values for hyperparameter $\lambda$}\label{fig6}
\end{figure}

\begin{figure*}[!t]%
\centering
\includegraphics[width= 1\linewidth]{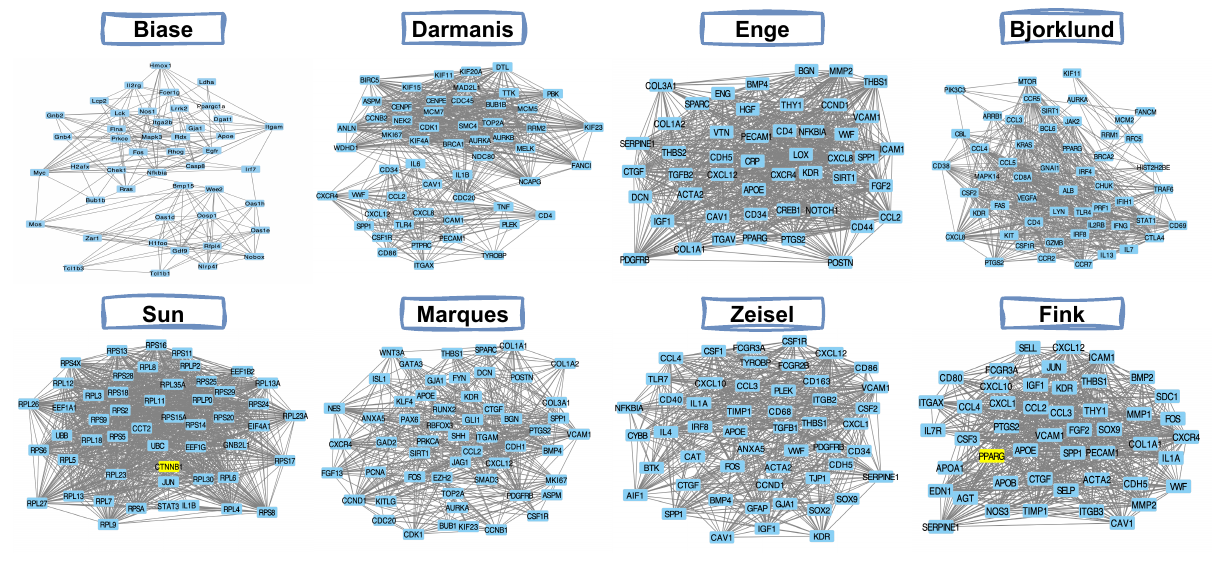}
\caption{The visualization of gene graphs pertaining to the eight benchmark datasets utilized in this study.}\label{fig7}
\end{figure*}

\subsection{Parameter Sensitivity Analysis.}

Our work underscores the considerable influence of dual-matrix alignment on clustering performance and introduces a weight coefficient in equation \ref{eq11}, represented by $\lambda$, to regulate the balance between cell loss and gene loss.  This section presents an in-depth analysis of the hyperparameter $\lambda$ and its impact on clustering outcomes.  To ascertain the optimal weight partitioning, we carried out experiments across four distinct datasets: Enge, Darmanis, Bjorklund, and Sun.  We tested various values of $\lambda$ in these experiments.  A smaller $\lambda$ value signifies a lesser weight attributed to cell loss, whereas a larger $\lambda$ value suggests a reduced weight for gene loss.  Figure \ref{fig6} displays the clustering performance of scEGA under varying $\lambda$ values.  The findings indicate that a balanced dual-matrix weight distribution is key to enhancing clustering embeddings, with a $\lambda$ value of 0.5 yielding consistently superior performance across all datasets.

\subsection{Exogenous Gene Visualization Analysis}

The core of this study lies in integrating exogenous gene association information into the clustering process, thereby enhancing the quality of cell embeddings.  In this section, to clearly demonstrate the gene-to-gene network, we visualized the gene graph.  Specifically, we entered gene sets corresponding to single-cell data into the renowned protein interaction network database, STRING, to acquire the gene adjacency tables.  These PPI networks depict the associative relationships among genes.  Subsequently, these tables were visualized using Cytoscape software, with the results displayed in Figure \ref{fig7}.  We present eight gene graphs derived from a variety of scRNA datasets.  This illustration substantiates the gene graph's existence, offering clear biological insights into gene features.  Such insights are vital for investigating gene functionality, regulation, and disease mechanisms.  Thus, maintaining the stability of gene features during the clustering process is of significant scientific relevance.

\section{Conclusion}
In conclusion, we have developed scEGA, an effective exogenous gene assisted clustering model for single-cell data, employing a dual-matrix alignment module to constrain cell and gene features. The dual-supervised optimization module enhances cluster embeddings while ensuring the stability of both cell and gene matrices during the optimization process. Our experimental findings show that the cell and gene graphs significantly improve embedding optimization, with our model surpassing other existing methods. 

In the future, we aim to investigate innovative approaches to random walks on the gene graph for more effective representations. Additionally, we plan to examine more similarity measures of cells to construct a more precise cell graph \cite{wang2023rnasmc,zhang2023review,lee2023benchmarking,kim2019impact,li2022compressed}. Collaborative training presents another exciting research direction, as we believe that integrating cell and gene data can mutually enhance clustering effectiveness \cite{nguyen2023embryosformer,ogier2023federated}.

\bibliographystyle{ieee}

\section*{Acknowledgments}
This work was supported in part by the National Key R\&D Program of China (no. 2020AAA0107100), and the National Natural Science Foundation of China (no. 62325604, 62276271).


\begin{IEEEbiography}[{\includegraphics[width=1in,height=1.10in,clip,keepaspectratio]{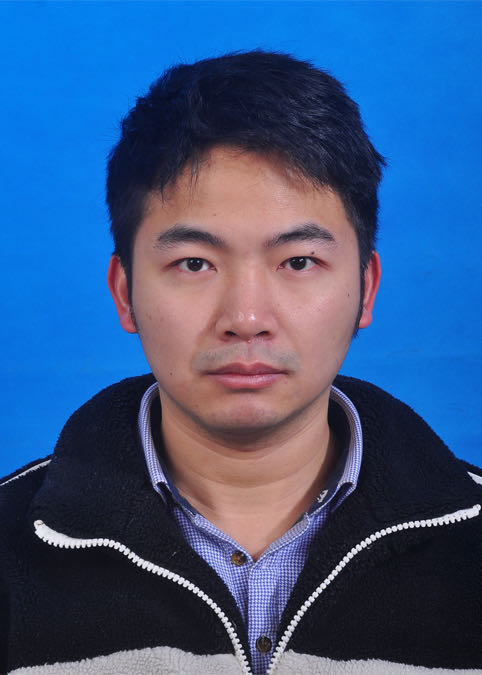}}]
{Dayu Hu} is currently pursuing a Ph.D. degree at the National University of Defense Technology (NUDT). Before joining NUDT, he got his BSc degree at Northeastern University (NEU). His current research interests include graph learning and bioinformatics. He has published several papers and served as PC member/ Reviewer in highly regarded journals and conferences such as ACM MM, AAAI, TNNLS, TKDE, TCBB, etc. 
\end{IEEEbiography}

\begin{IEEEbiography}[{\includegraphics[width=1in,height=1.10in,clip,keepaspectratio]{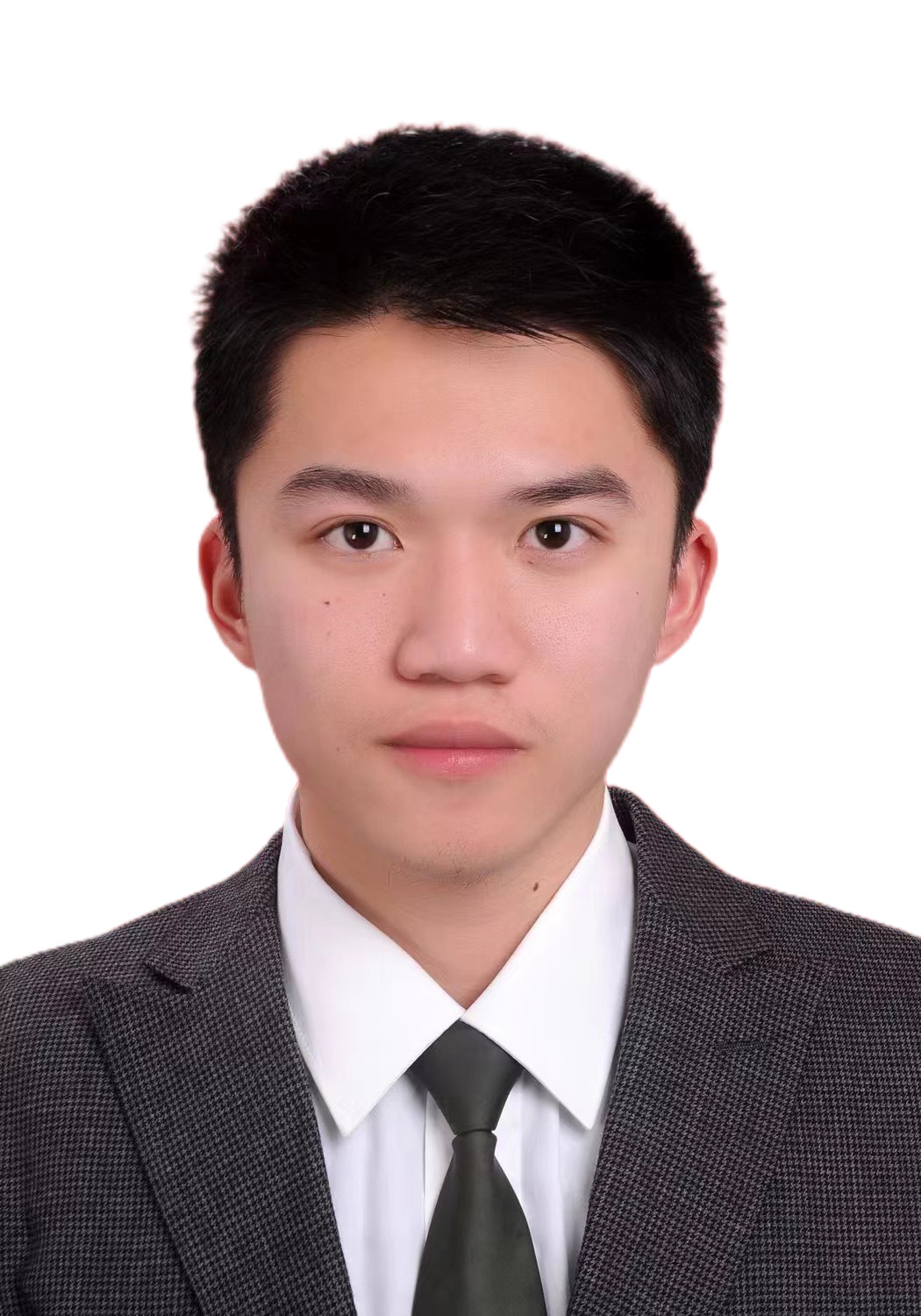}}]
{Ke Liang} is currently pursuing a Ph.D. degree at the National University of Defense Technology (NUDT). Before joining NUDT, he got his BSc degree at Beihang University (BUAA) and received his MSc degree from the Pennsylvania State University (PSU). His current research interests include knowledge graphs, graph learning, and healthcare AI. He has published several papers in highly regarded journals and conferences such as SIGIR, AAAI, ICML, ACM MM, IEEE TNNLS, IEEE TKDE, etc.
\end{IEEEbiography}

\begin{IEEEbiography}
[{\includegraphics[width=0.8in,clip,keepaspectratio]{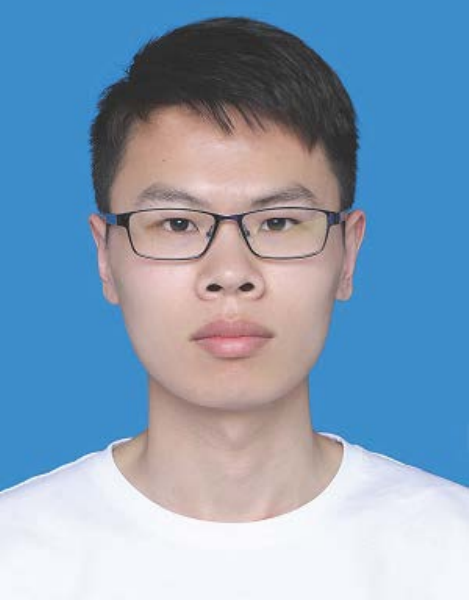}}]
{Hao Yu} is presently pursuing a Ph.D. degree at the National University of Defense Technology (NUDT).
He earned a B.Eng degree in computer science from Inner Mongolia University, Hohhot, China, in 2019, and, subsequently, in 2022, obtained an MA.Sc in cyberspace science and technology from Beijing Institute of Technology, Beijing, China.
His current research focuses on AI security and Federated Learning.
He has authored several papers in top-level journals and conferences, such as IEEE TIFS, TDSC, and ACM MM, and served as a Reviewer for highly regarded journals, such as IEEE TIFS, IEEE TKDE, ACM TOIS, etc.
\end{IEEEbiography}

\begin{IEEEbiography}[{\includegraphics[width=1in,height=1.10in,clip,keepaspectratio]{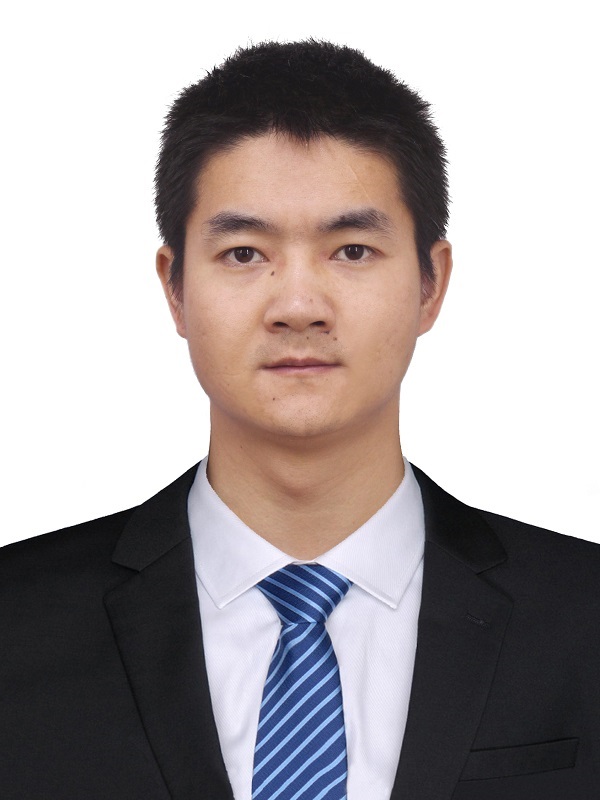}}]{Xinwang Liu} received his PhD degree from National University of Defense Technology (NUDT), China. He is now Professor of School of Computer, NUDT. His current research interests include kernel learning and unsupervised feature learning. Dr. Liu has published 60+ peer-reviewed papers, including those in highly regarded journals and conferences such as IEEE T-PAMI, IEEE T-KDE, IEEE T-IP, IEEE T-NNLS, IEEE T-MM, IEEE T-IFS, ICML, NeurIPS, ICCV, CVPR, AAAI, IJCAI, etc. He serves as the associated editor of TNNLS, TCYB and Information Fusion Journal. More information can be found at {https://xinwangliu.github.io/}.
\end{IEEEbiography}

\end{document}